\documentclass[lettersize,journal]{IEEEtran}
\usepackage{amsmath,amsfonts}
\usepackage{algorithmic}
\usepackage{algorithm}
\usepackage{array}
\ifCLASSOPTIONcompsoc
	\usepackage[caption=false, font=normalsize, labelfont=sf, textfont=sf]{subfig}
\else
	\usepackage[caption=false, font=footnotesize]{subfig}
\usepackage{textcomp}
\usepackage{multirow}
\usepackage{stfloats}
\usepackage{url}
\usepackage{verbatim}
\usepackage{graphicx}
\usepackage{cite}
\usepackage{float}
\usepackage{pifont}
\usepackage{orcidlink}
\usepackage{color}
\begin{document}

\title{SCAResNet: A ResNet Variant Optimized for Tiny Object Detection in Transmission and Distribution Towers}

\author{Weile Li~\orcidlink{0009-0007-8222-6917},~\IEEEmembership{Student Member,~IEEE,} Muqing Shi~\orcidlink{0009-0004-0411-5208},~\IEEEmembership{Student Member,~IEEE,} \\ Zhonghua Hong~\orcidlink{0000-0003-0045-1066},~\IEEEmembership{Member,~IEEE.}
        % <-this % stops a space
\thanks{Manuscript received 23 August 2023; accepted 11 September 2023. Date of publication 14 September 2023; date of current version 12 September 2023. This work was supported in part by the National Natural Science Foundation of China under Grant 42241164. \textit{(W. Li and M. Shi are co-first authors.)(Corresponding author: Zhonghua Hong.)}}% <-this % stops a space
\thanks{The authors are with the College of Information Technology, Shanghai Ocean University, Shanghai 201306, China. (e-mail: 3493617871lisavila@gmail.com, shimuqing0309@gmail.com, zhhong@shou.edu.cn)}}

% The paper headers
\markboth{IEEE GEOSCIENCE AND REMOTE SENSING LETTERS,~Vol.~20,~2023}%
{Shell \MakeLowercase{\textit{et al.}}: SCAResNet: A ResNet Variant Optimized for Tiny Object Detection in Transmission and Distribution Towers}

\maketitle

\begin{abstract}
Traditional deep learning-based object detection networks often resize images during the data preprocessing stage to achieve a uniform size and scale in the feature map. Resizing is done to facilitate model propagation and fully connected classification. However, resizing inevitably leads to object deformation and loss of valuable information in the images. This drawback becomes particularly pronounced for tiny objects like distribution towers with linear shapes and few pixels. To address this issue, we propose abandoning the resizing operation. Instead, we introduce Positional-Encoding Multi-head Criss-Cross Attention. This allows the model to capture contextual information and learn from multiple representation subspaces, effectively enriching the semantics of distribution towers. Additionally, we enhance Spatial Pyramid Pooling by reshaping three pooled feature maps into a new unified one while also reducing the computational burden. This approach allows images of different sizes and scales to generate feature maps with uniform dimensions and can be employed in feature map propagation. Our SCAResNet incorporates these aforementioned improvements into the backbone network ResNet. We evaluated our SCAResNet using the Electric Transmission and Distribution Infrastructure Imagery dataset from Duke University. Without any additional tricks, we employed various object detection models with Gaussian Receptive Field based Label Assignment as the baseline. When incorporating the SCAResNet into the baseline model, we achieved a 2.1\% improvement in mAP\textsubscript{s}. This demonstrates the advantages of our SCAResNet in detecting transmission and distribution towers and its value in tiny object detection. The source code is available at \url{https://github.com/LisavilaLee/SCAResNet_mmdet}.
\end{abstract}


\section{Introduction}
\IEEEPARstart{D}{etecting} transmission and distribution towers is crucial for the safe and reliable operation of the power grid, as the location and quantity of these towers are vital parameters in designing the topology of the electrical network and planning for its expansion. Combining remote sensing (RS) and deep learning techniques as a widely adopted method for object detection has several advantages, including the ability to quickly cover large areas, reduce human errors, and improve detection accuracy. Convolutional neural networks (CNNs) have become the primary approach in deep learning because they can automatically learn discriminative features from raw image pixels, capturing complex spatial relationships between image regions and achieving better detection performance. Mainstream CNN-based object detectors can be divided into anchor-based and anchor-free categories. Anchor-based detectors use predefined anchor boxes to predict object position and size, which offer better accuracy but may be influenced by anchor box design biases. Anchor-free detectors directly regress the object's center and size without using anchor boxes.

\begin{figure}
\centering
\begin{minipage}{.40\columnwidth}
    \subfloat[]{
      \label{a} 
      \includegraphics[width=.79\columnwidth]{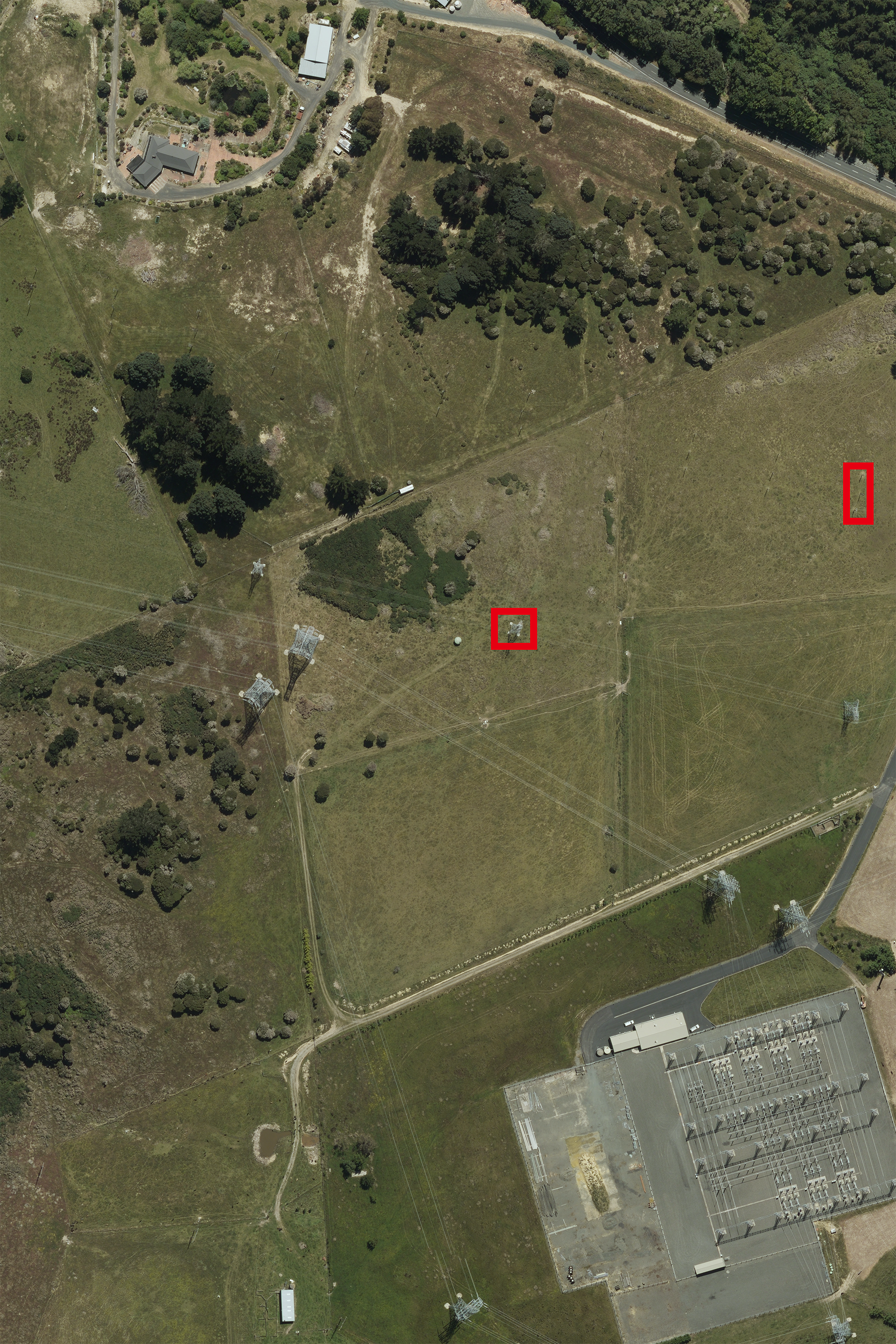}}
\end{minipage}
\begin{minipage}{.25\columnwidth}
    \subfloat[]{
      \label{b} 
      \includegraphics[width=.90\columnwidth]{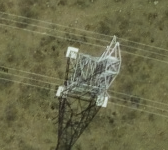}}\\
    \subfloat[]{
      \label{c} 
      \includegraphics[width=.90\columnwidth]{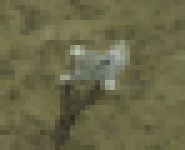}}
\end{minipage}
\begin{minipage}{.25\columnwidth}
    \subfloat[]{
      \label{d} 
      \includegraphics[width=.34\columnwidth]{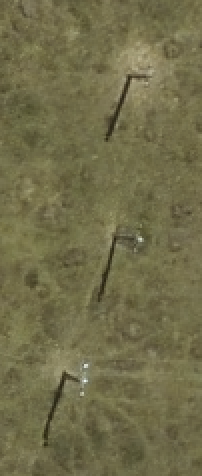}}\\
    \subfloat[]{
      \label{e} 
      \includegraphics[width=.34\columnwidth]{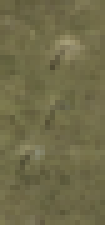}}
\end{minipage}
\caption{(a) is an original picture with dimensions $5760\times 3840$ pixels. (b) and (d) represent a transmission tower and three distribution towers, respectively, extracted from (a). (c) and (e) show the resized versions of (b) and (d) obtained by resizing (a) to $800\times 800$ pixels.}
\label{dataset}
\vspace{-2ex}
\end{figure}

Nevertheless, detecting transmission and distribution towers is challenging due to their tiny size and lack of distinct features, especially since distribution towers are usually black linear objects in remote sensing images. Tiny objects typically have a smaller spatial footprint and lower contrast with their surroundings, which makes them harder to detect. Moreover, existing object detectors tend to favor larger objects when assigning positive and negative samples\cite{todassign}, leading to most of the tiny targets not getting a positive sample assignment\cite{todassign}. Therefore, the detection performance is significantly reduced when detecting tiny objects, such as transmission towers and especially distribution towers. 

Much work today is dedicated to solving the difficulties of tiny object detection. Multi-scale feature learning involves learning objects of varying sizes separately, mainly solving the problem of less discriminative features of tiny objects. Feature pyramid-based methods\cite{fpn} mainly use low-level spatial and high-level semantic information to enhance object features. To address the issue of tiny objects not being assigned to samples, Xu et al. proposed Gaussian Receptive Field based Label Assignment (RFLA)\cite{rfla}. As a label assignment designed explicitly for tiny objects, RFLA draws inspiration from Normalized Gaussian Wasserstein Distance\cite{nwd} and Effective Receptive Fields\cite{erf}, employing Gaussian modeling of the ground-truth boxes and receptive fields to re-measure their distance for label assignment. This approach improves label assignment probability for tiny objects and yields good results for both anchor-based and anchor-free detectors.

However, the current related work pays limited attention to the impact of data preprocessing on tiny object detection. During the data preprocessing stage, conventional practice is to perform a resizing operation on the data before it enters the object detection network, aiming to achieve a uniform size and scale for images of different sizes and scales. This practice aims to reduce the computational burden on the model and, more importantly, ensure that feature maps can produce feature vectors of the same length. This is an indispensable prerequisite for subsequent fully connected classification. Unfortunately, for tiny objects, due to their smaller number of pixels and relatively lower richness in features, the loss of pixels during resizing in data preprocessing has a more severe impact on extracting features from tiny objects compared to larger objects. As shown in Fig. \ref{dataset}, compared to the resized image of a transmission tower with rich features, the resized image of a distribution tower with a tiny, black linear shape loses more distinctive features, even becoming a blurry black dot. This is highly disadvantageous for subsequent feature extraction, as these tiny objects lose valuable features from the outset\cite{highres}.

Therefore, we proposed abandoning the traditional data preprocessing step of resizing. Instead, in the propagation of the backbone network, we initially perform Positional-Encoding Multi-Head Criss-Cross Attention on feature maps of varying sizes and scales. This allows the model to learn about the overall contextual information across the entire image in multiple sub-representation spaces so that the model can learn richer feature information about the distribution tower's position and its surrounding environment in remote sensing images. Subsequently, three rounds of pooling will be applied to feature maps of different sizes and scales. The flattened feature vectors of the pooled feature maps will be reshaped into new feature maps with the same size and scale. Our designed backbone network, SCAResNet, which incorporates the aforementioned innovative modules into ResNet\cite{resnet}, achieves promising results on the Electric Transmission and Distribution Infrastructure Imagery dataset\cite{etdii} released by Duke University.

\section{Methods}

\begin{figure}[h]
	\centering
	\includegraphics[scale=0.20]{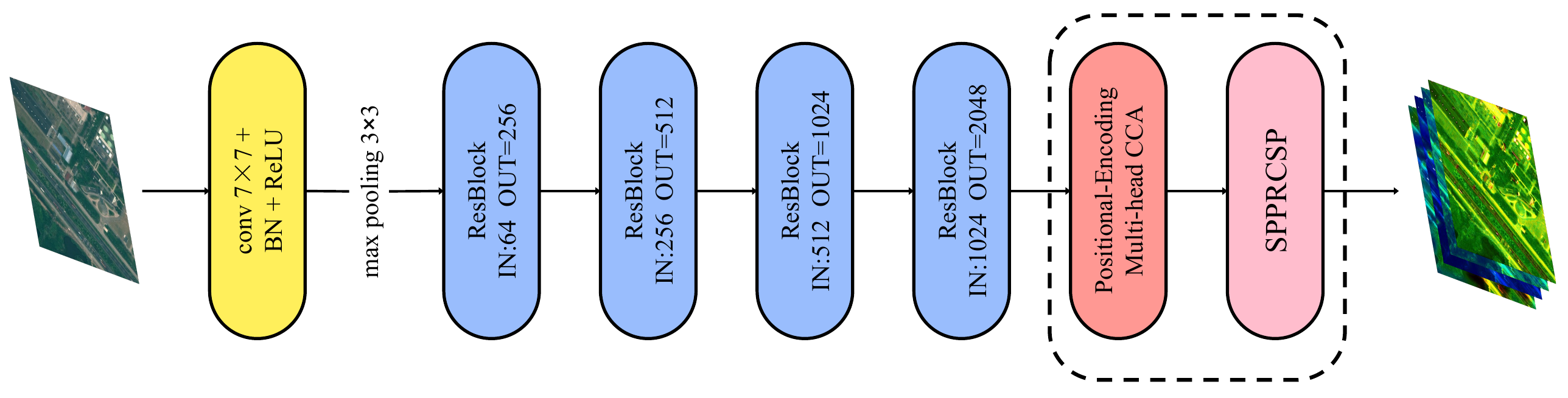}
	\vspace{-2ex}
	\caption{Overview of our SCAResNet. The boxed section represents the modifications we made to improve ResNet.}
	\label{pipeline}
\end{figure}

SCAResNet combines the Positional-Encoding Multi-head Criss-Cross Attention (Positional-Encoding Multi-head CCA) and SPPRCSP modules based on ResNet, and the architecture is shown in Fig. \ref{pipeline}.

\subsection{Positional-Encoding Multi-head CCA Module}

\begin{figure}[h]
	\centering
	\includegraphics[scale=0.20]{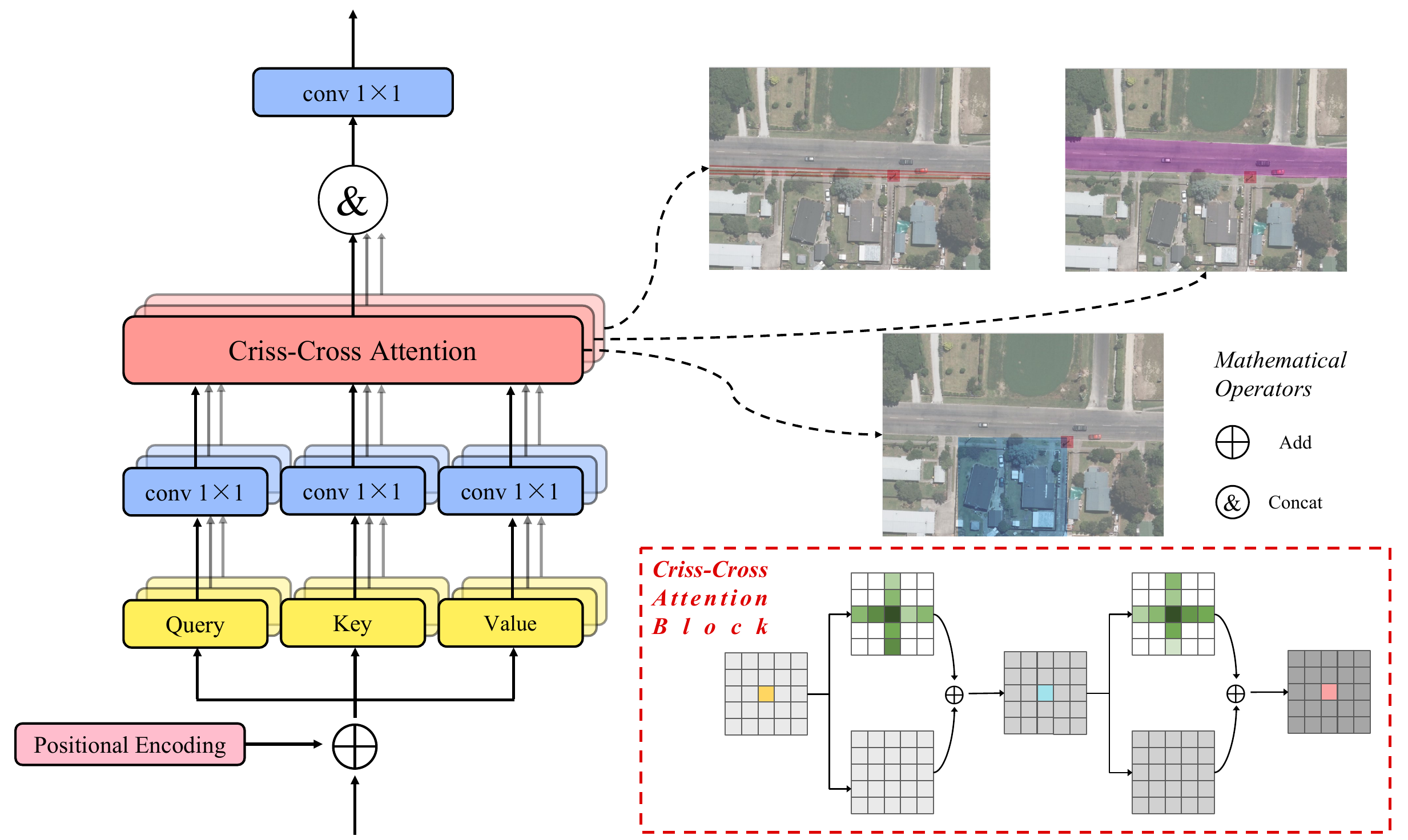}
	\vspace{-2ex}
	\caption{Overview of Positional-Encoding Multi-head CCA Module.}
	\label{Multi-head-CCA}
\end{figure}

One challenge in detecting transmission and distribution towers is the limited availability of rich features, especially for distribution towers that often exhibit tiny black linear shapes. To alleviate the problem of less distinctive features in distribution towers, we employ Criss-Cross Attention (CCA)\cite{cca} improved based on Self-Attention\cite{attentionall} to capture the contextual information and correlations of distribution towers from long-term dependency relationships, resulting in towers with richer semantics. As illustrated by the Criss-Cross Attention Block module in Fig. \ref{Multi-head-CCA}, CCA replaces the full-map correlation matrix transformation in Self Attention with two consecutive row-column correlation matrix transformations, significantly reducing the parameter count required by Self Attention while maintaining high accuracy.

However, single-head attention processes input data in a single representation, which may limit the model's ability to capture the full complexity of the data. Multi-head attention, in contrast, projects input data into multiple representation subspaces, enabling the model to capture diverse aspects of information and integrate them for the final output\cite{attentionall}.Therefore, Multi-head CCA can allocate multiple attention focuses to different points of interest, such as the surrounding environment and the global context, as shown in Fig. \ref{Multi-head-CCA}, including power lines, residential blocks, and roads in the image. This comprehensive learning approach aids in determining which environmental features are more likely associated with the presence of a distribution tower, thereby assisting in detection.

When computing attention, the influence of different positions' keys was not considered, so the purpose of preprocessing with absolute positional encoding is to add positional information to capture positional relationships in the attention mechanism. We use a positional encoding method based on sine and cosine functions, which encodes positional information as periodic high-dimensional vectors that can be easily added to the input data.

\subsection{SPPRCSP Module}
\begin{figure*}[t]
	\centering
	\includegraphics[scale=0.30]{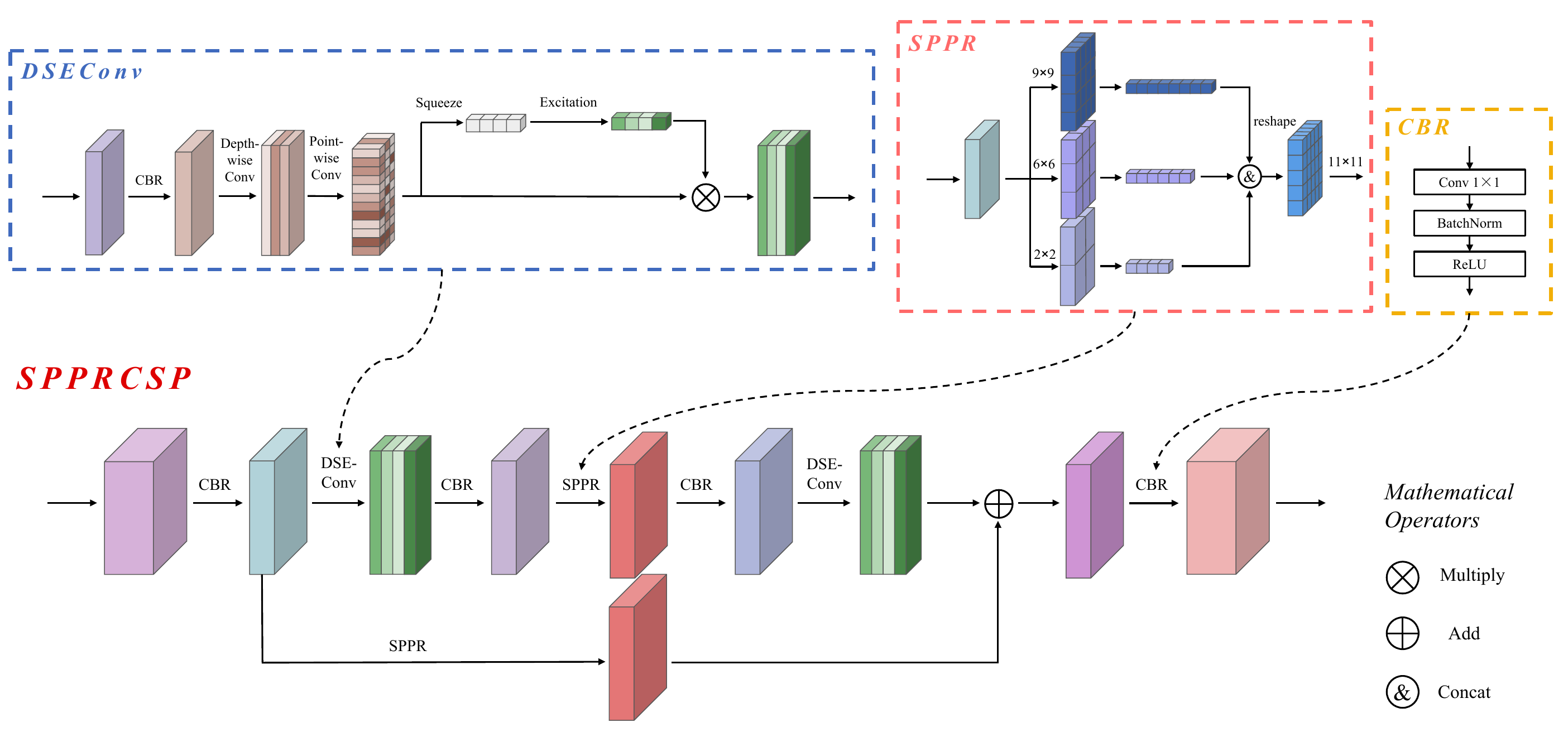}
	\vspace{-2ex}
	\caption{Overview of SPPRCSP.}
	\label{spprcsp}
\end{figure*}

After undergoing comprehensive feature extraction of contextual information related to distribution towers with preserved information by the Positional-Encoding Multi-head CCA module, feature maps of varying sizes and scales will go through our designed Spatial Pyramid Pooling Reshaping Cross Stage Partial (SPPRCSP) module. This module ensures the output of feature maps with a uniform size and scale while avoiding loss of accuracy and reducing computational costs for subsequent tasks such as fully connected classification and others.

Spatial Pyramid Pooling (SPP)\cite{spp}, proposed by He et al., divides feature maps of various sizes and scales into a set (usually 3) of pyramid layers with fixed-size grids. The features within each grid region are independently pooled. Ultimately, the pooled features from each grid region of pyramid layers are concatenated to form a fixed-length feature vector. However, this feature vector cannot propagate within a network organized in terms of feature maps; it can only serve as input to the classification network. Thus, we introduce the Spatial Pyramid Pooling Reshaping (SPPR) layer to enable feature vector propagation within the network.

As illustrated by the SPPR module in Fig. \ref{spprcsp}, in the early stages of the SPPR layer, the steps of the SPP layer are maintained unchanged, with the difference being the reshaping of the concatenated feature vectors at the end. Unlike the standard SPP layer, in order to successfully reshape the concatenated feature vectors into a feature map, the levels of the pooled feature maps in the SPPR layer cannot be arbitrarily specified; they must satisfy the following equations:
\begin{equation} 
\begin{cases}
	{\;x^{2}+y^{2}+z^{2}\,=\,w^{2}} \\
	{\;(x,\,y,\,z,\,w)\,=\,(2a,\,2b,\,2c-1,\,2c+2d-1)} \\
	{\;a^{2}+b^{2}\,=\,d\cdot(2c+d-1)} \\
\end{cases}
\label{eq:level}
\end{equation}

where ${x}$, ${y}$, ${z}$, and ${w}$ are the levels of the three pooled feature maps and the reshaped feature map. ${a}$, ${b}$, ${c}$, and ${d}$ are arbitrary positive integers. In our research, ${x}$, ${y}$, ${z}$, and ${w}$ are determined to be ${9}$, ${6}$, ${2}$, and ${11}$, respectively.

With the pooled feature map levels established, we can deduce the parameters for the pooling layer, including the pooling kernel, stride, and padding values. In the original SPP paper's equations, during the pooling process of the feature map, it is possible for the padding size to exceed half the size of the pooling kernel. This situation can lead to the truncation of edge information and the loss of valuable features, which in turn hampers the model training process. Consequently, we present the revised equations for calculating the parameters of SPPR pooling layers as follows:
\begin{equation}
\begin{aligned} 
	t\,=\,\lfloor\frac{l}{h}\rfloor+(h\bmod l)+1 \label{eq:calcult}
\end{aligned}
\end{equation}

\begin{equation}
\begin{aligned}
	(t>l)\cup((t\,=\,l)\cap (\frac{h}{l-1}\bmod 2\,=\,0)) \label{eq:jdgt} 
\end{aligned}
\end{equation}

where $t$ is a judgment value calculated using Equation (\ref{eq:calcult}). $h$ represents the height or width of the input feature map. $l$ represents the level of the pooled feature map.

If Equation (\ref{eq:jdgt}) is satisfied, then:
\begin{equation} 
\begin{aligned}
\begin{cases}
	\;stride = \lceil \frac{h}{l} \rceil \\ 
	\;kernel= \lceil \frac{h}{l} \rceil \\ 
	\;padding = \lceil\frac{l\cdot kernel - h}{2}\rceil
\end{cases}
\end{aligned}
\label{eq:truet}
\end{equation}

If Equation (\ref{eq:jdgt}) is not satisfied, then:
\begin{equation} 
\begin{cases}
	\; stride = \lfloor \frac{h}{l} \rfloor \\ 
	\; kernel= h-(l-1)\cdot stride \\ 
	\; padding = 0
\end{cases}
\label{eq:falset}
\end{equation}

Due to the absence of resizing, the feature map sizes from high-resolution remote sensing images are larger. This, in turn, imposes a significant computational burden on the training process. To address this, we introduce a Cross Stage Partial (CSP)\cite{csp} structure and DSEConv module. As depicted in Fig. \ref{spprcsp}, the feature map entering the SPPRCSP structure undergoes channel compression to reduce parameters. To compensate for the accuracy loss caused by channel compression and feature maps' pooling\cite{pooling1}\cite{pooling2}, the feature map integrates detailed information into the results of the SPPR layer through skip connections. Additionally, we replace ordinary convolutions with depth-wise and point-wise convolutions to reduce parameters and FLOPs\cite{dsc} while introducing Squeeze and Excitation Attention\cite{sea} to aid in selecting channels more conducive to tiny object detection from a global perspective.

\begin{table*}[htbp]
  \centering
%  \captionsetup{labelformat=simple}
  \caption{Ablation Experiment on ETDII Dataset}
  \resizebox{\textwidth}{!}{
    \begin{tabular}{c c c c l l l l l l}
      \hline
      Num & Baseline(RFLA) & PE Multi-head CCA & SPPRCSP & mAP(\%) & mAP\textsubscript{50}(\%) & mAP\textsubscript{s}(\%) & mAP\textsubscript{m}(\%) & mAP\textsubscript{l}(\%)\\
      \hline
      1 & \ding{51} & & & 22.0 & 48.8 & 5.5 & 19.4 & \underline{36.2} \\
      2 & \ding{51} & \ding{51} & & 22.7 & 50.2 & 5.7 & 20.4 & \textbf{37.9} \\
      3 & \ding{51} & & \ding{51} & \textbf{23.7} & \underline{51.7} & \underline{6.6} & \textbf{21.3} & 32.8 \\
      4 & \ding{51} & \ding{51} & \ding{51} & \underline{23.3} (+1.3) & \textbf{53.1} (+4.3) & \textbf{7.6} (+2.1) & \underline{20.8} (+1.4)& 34.4 (-1.8) \\
      \hline
    \end{tabular}
  }
  \label{ablation}
\end{table*}

\begin{table*}[htbp]
  \centering
  \caption{Performance of Different Models with SCAResNet.}
  \resizebox{\textwidth}{!}{
    \begin{tabular}{c l l l l l l}
      \hline
      Methods & mAP(\%) & mAP\textsubscript{50}(\%) & mAP\textsubscript{s}(\%) & mAP\textsubscript{m}(\%) & mAP\textsubscript{l}(\%)\\
      \hline
      Cascade R-CNN w/ RFLA & 29.0 & 61.0 & 13.3 & 22.1 & \textbf{45.3} \\
      Faster R-CNN w/ RFLA & 27.8 & 59.3 & 11.7 & 21.7 & \textbf{45.4} \\
      FCOS w/ RFLA & 22.0 & 48.8 & 5.5 & 19.4 & \textbf{36.2} \\
      \hline
      Cascade R-CNN w/ RFLA w/ SCAResNet & \textbf{29.8} (+0.8) & \textbf{62.6} (+1.6) & \textbf{14.6} (+1.3) & \textbf{23.0} (+0.9) & 43.2 (-2.1) \\
      Faster R-CNN w/ RFLA w/ SCAResNet & \textbf{28.1} (+0.3) & \textbf{61.6} (+2.3) & \textbf{12.9} (+1.2) & \textbf{22.5} (+0.8) & 41.7 (-3.7) \\
      FCOS w/ RFLA w/ SCAResNet & \textbf{23.3} (+1.3) & \textbf{53.1} (+4.3) & \textbf{7.6} (+2.1) & \textbf{20.8} (+1.4) & 34.4 (-1.8)\\
	  \hline
    \end{tabular}
  }
  \label{models}
\end{table*}

\section{Experiment Results}
\subsection{Dataset}
We conducted experiments using the Electric Transmission and Distribution Infrastructure Imagery (ETDII) dataset, which is a publicly available dataset from Duke University. The dataset was sourced from various providers, including CT ECO, USGS, LINZ, and SpaceNet. It comprises $494$ image tiles from six countries, namely the USA, Sudan, New Zealand, Mexico, China, and Brazil, each with a unique terrain type and a resolution of $0.3$m. The detection categories are Distribution Tower (DT) and Transmission Tower (TT), with $16,418$ and $1,385$ ground truth boxes for DT and TT, respectively. According to the size classification criteria of the standard COCO dataset\cite{coco}, objects with a size less than or equal to $32\times 32$ pixels are considered small, objects with a size greater than $32\times 32$ pixels but less than or equal to $96\times 96$ pixels are considered medium, and objects with a size greater than $96\times 96$ pixels are considered large. In ETDII dataset, there are $12,713$ small objects, out of which $6,342$ are smaller than or equal to $20\times 20$ pixels; there are $4,723$ medium objects; and there are $367$ large objects.

\subsection{Experiment Settings}
All experiments were conducted on a computer with one NVIDIA GeForce RTX 3060 Laptop GPU. Model training was based on PyTorch\cite{pytorch}, and the core code was built on MMdetection\cite{mmdetection}. A pre-trained model from ImageNet\cite{imagenet} was used. All models were trained using a stochastic gradient descent (SGD) optimizer with $0.9$ momentum, $0.0001$ weight decay, and $2$ batch sizes for $30$ epochs. The initial learning rate was set to $0.005$ and decayed $0.001$ at the $20$th and $27$th epochs. We conduct experiments using a SCAResNet with $50$ layers. All other parameter settings were the same as in RFLA-based detectors.

\subsection {Ablation Study}
We evaluated the effect of different modules on SCAResNet performance by ablation experiment to verify the effectiveness of each module. The results of the ablation experiments are shown in Table \ref{ablation}.

\subsubsection {Effect of Positional-Encoding Multi-head CCA}
In our experiments, the RFLA-based detectors with the Positional-Encoding Multi-head CCA module showed a $0.7$\% improvement in mAP, $1.4$\% in mAP\textsubscript{50}, $0.2$\% in mAP\textsubscript{s}, $1.0$\% in mAP\textsubscript{m}, and $1.7$\% in mAP\textsubscript{l} compared to RFLA-based detectors alone.

\subsubsection {Effect of SPPRCSP}
In our experiments, the RFLA-based detectors with the SPPRCSP module showed a $1.7$\% improvement in mAP, $2.5$\% in mAP\textsubscript{50}, $1.1$\% in mAP\textsubscript{s}, and $1.9$\% in mAP\textsubscript{m}, but a $3.4$\% decrease in mAP\textsubscript{l} compared to the RFLA-based detectors alone.

\subsubsection {Effect of Combining Positional-Encoding Multi-head CCA and SPPRCSP}
The RFLA-based detectors with both Positional-Encoding Multi-head CCA and SPPRCSP modules combine the advantages of both modules and show a $1.3$\% improvement in mAP, $4.3$\% in mAP\textsubscript{50}, $2.1$\% in mAP\textsubscript{s}, and $1.4$\% in mAP\textsubscript{m} compared to the RFLA-based detectors alone.

\subsection {Contrast Experiments}
\begin{figure}
\centering
	\includegraphics[scale=0.35]{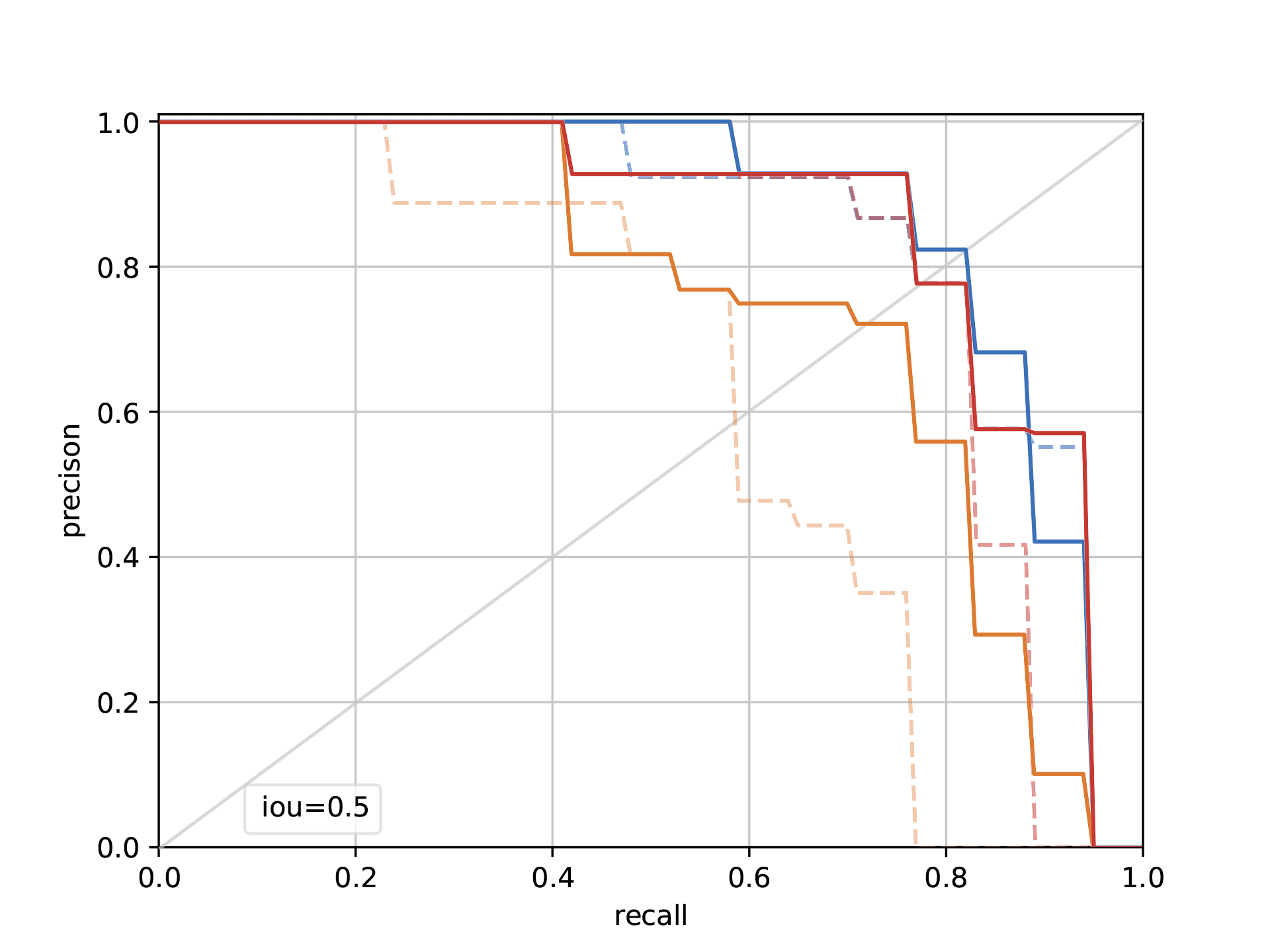}
	\caption{Precision-Recall Curves of Contrast Experiments. The dashed line represents the baseline, while the solid line represents the detector using SCAResNet. The blue, red, and orange colors correspond to Cascade R-CNN, Faster R-CNN, and FCOS, respectively.}
	\vspace{-2ex}
    \label{pr}
\end{figure}

In order to validate the performance of SCAResNet in detecting transmission and distribution towers across multiple detector architectures, we conducted contrast experiments on the ETDII dataset by replacing the backbone network of Cascade R-CNN\cite{cascadercnn}, Faster R-CNN\cite{fasterrcnn}, FCOS\cite{fcos} with our designed SCAResNet. Table \ref{models} shows the experimental results, and Fig. \ref{pr} illustrates the precision-recall performance of each detector.

As can be seen, our SCAResNet-based detectors show varying degrees of improvement in mAP, mAP\textsubscript{50}, mAP\textsubscript{s}, and mAP\textsubscript{m} when applied to different detectors. However, there is a slight decrease in mAP\textsubscript{l}. This is due to the imbalance caused by the much smaller number of large objects compared to medium and small ones.

\begin{figure}[t]
    \centering
	\captionsetup{font=tiny}
    \subfloat[FCOS Baseline]{\includegraphics[scale=0.06]{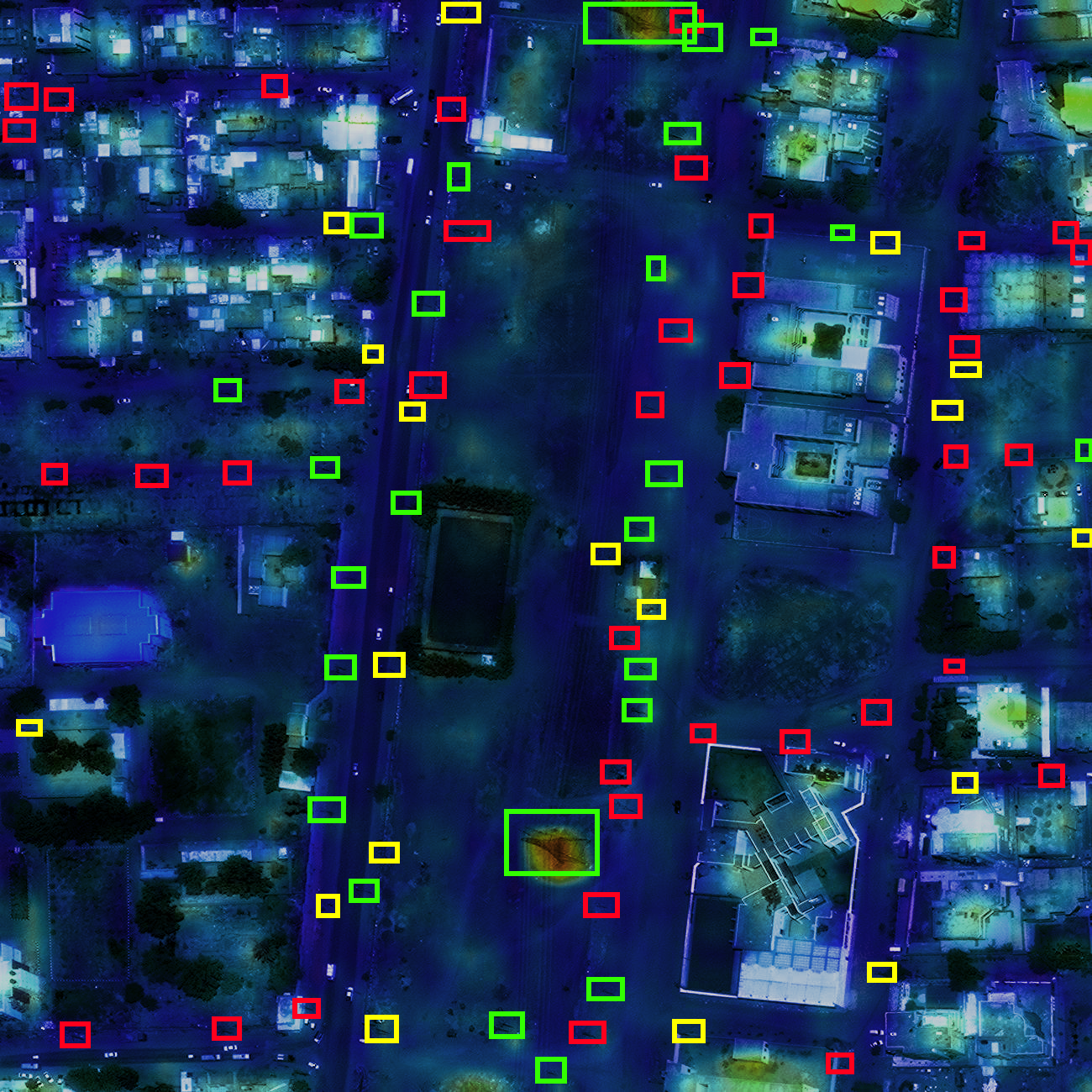}}%
    \label{FB}
    \subfloat[Faster R-CNN Baseline]{\includegraphics[scale=0.06]{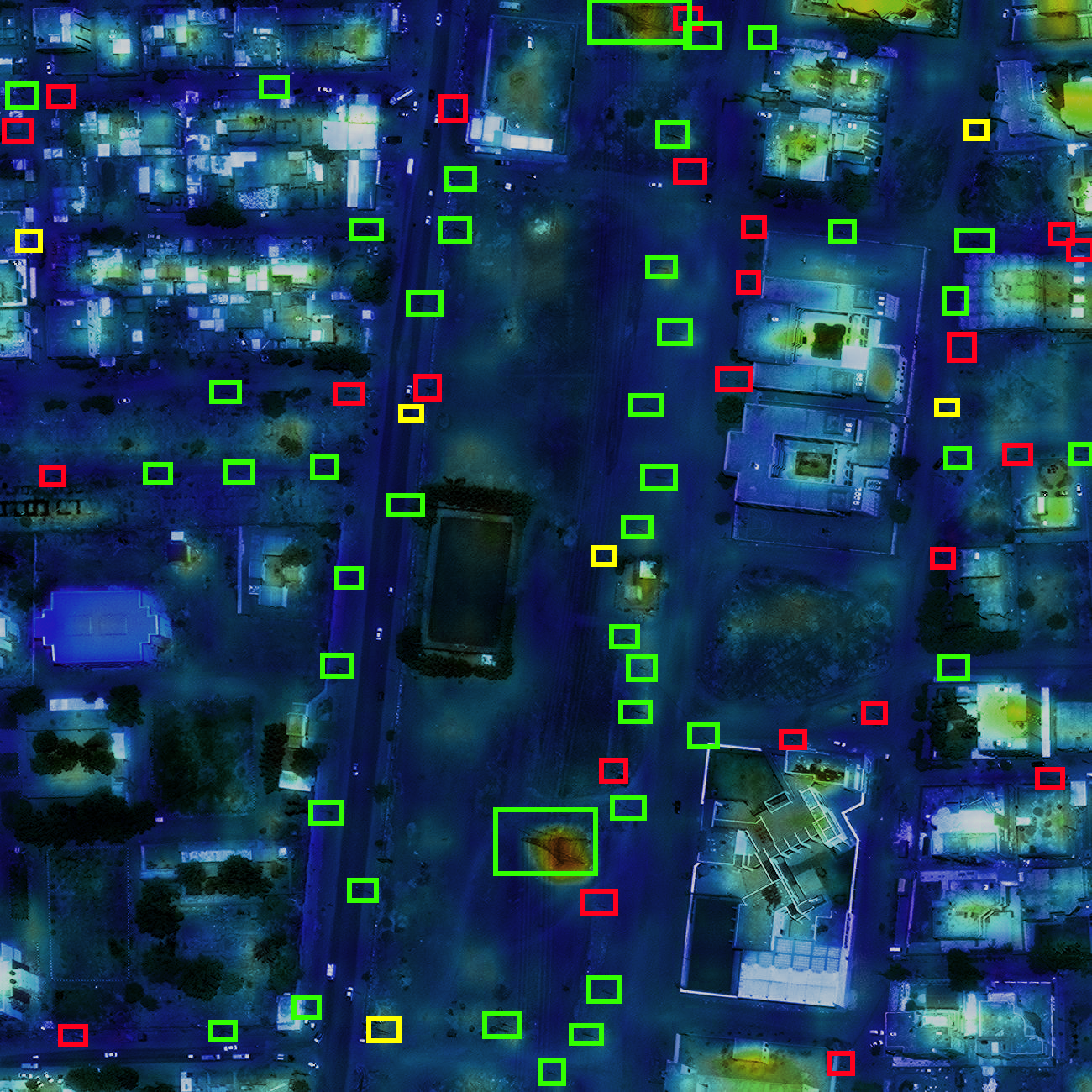}}%
    \label{FRB}    
	\subfloat[Cascade R-CNN Baseline]{\includegraphics[scale=0.06]{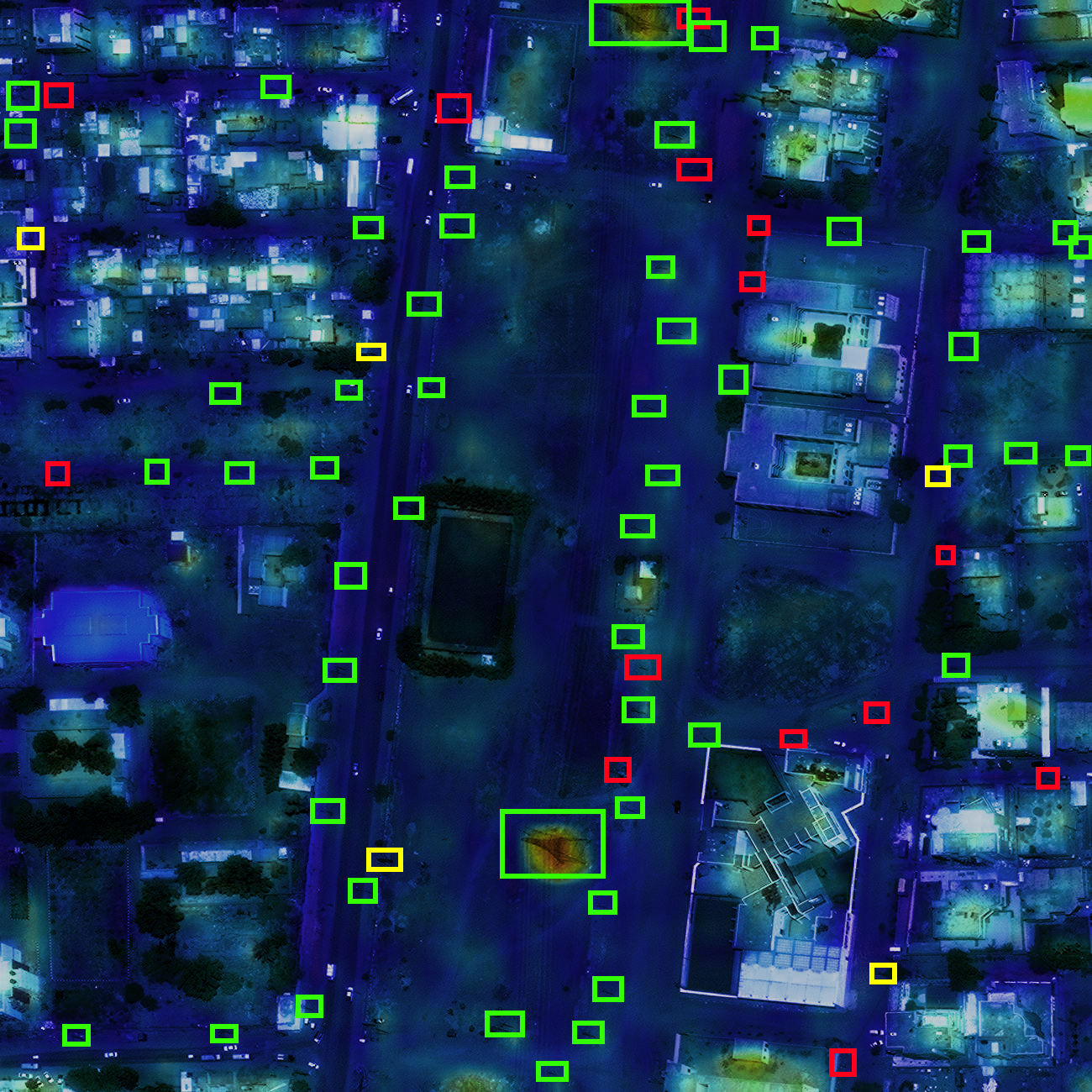}}%
    \label{CB}\\
    \subfloat[FCOS Baseline w/ SCAResNet]{\includegraphics[scale=0.06]{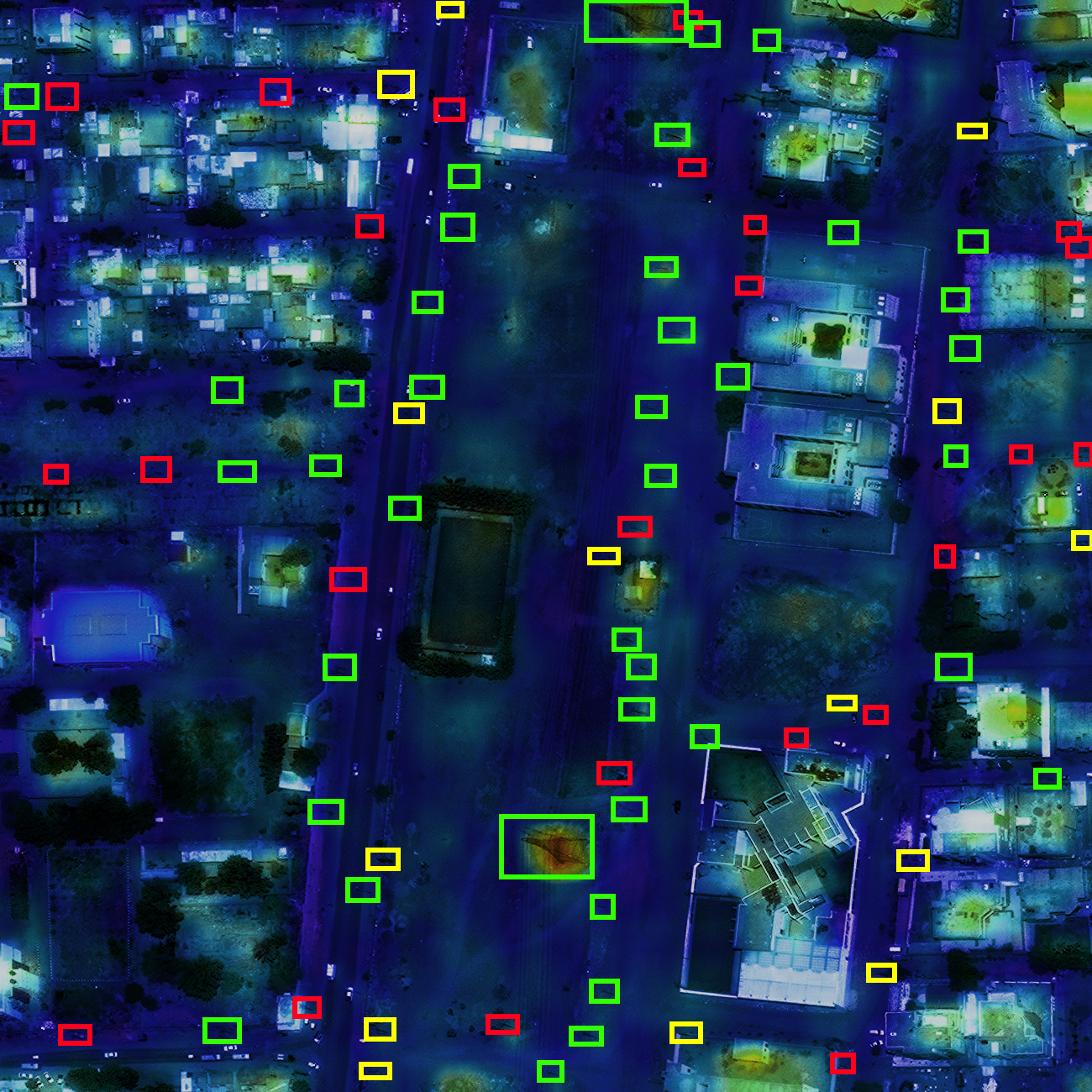}}%
    \label{FS}
    \subfloat[Faster R-CNN Baseline w/ SCAResNet]{\includegraphics[scale=0.06]{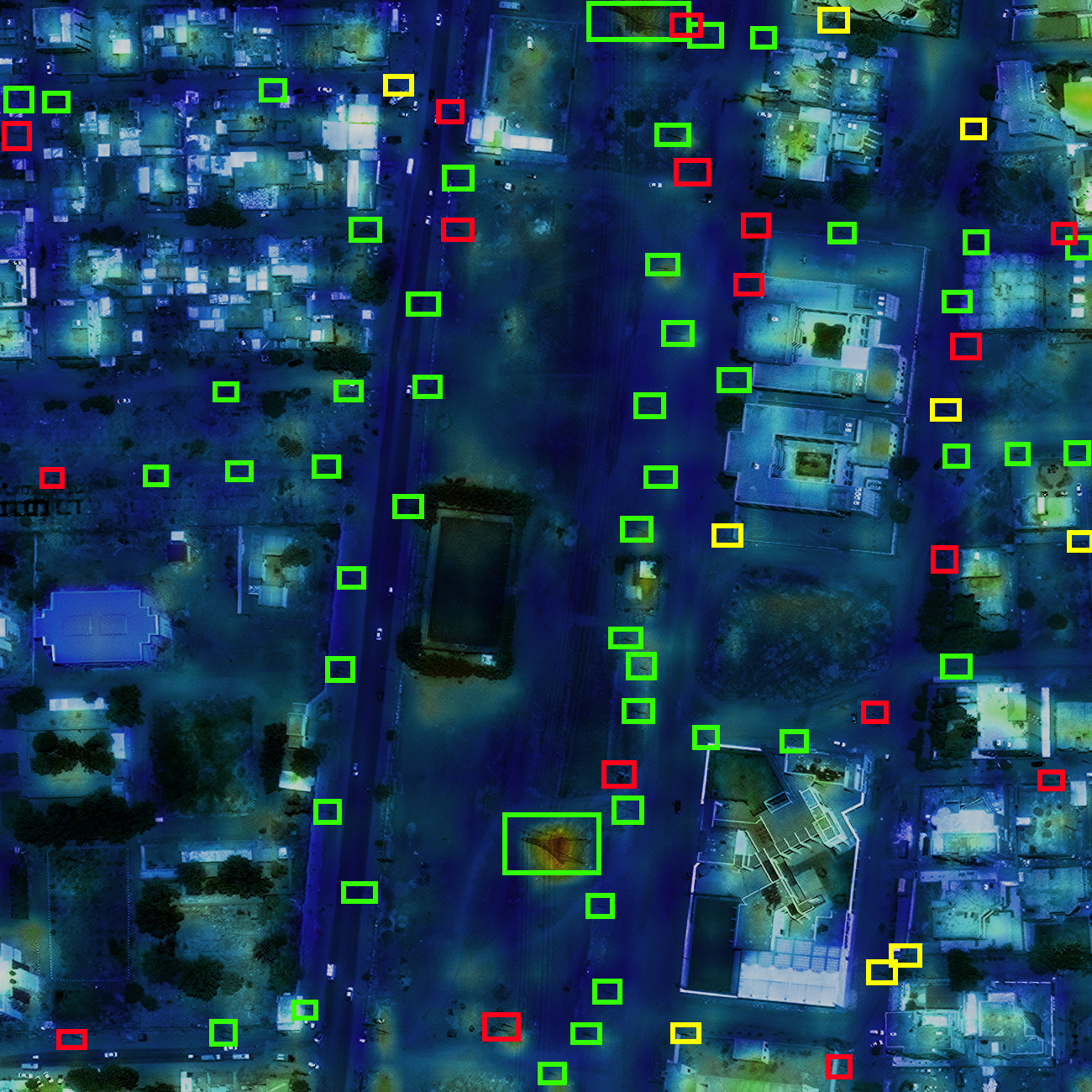}}%
    \label{FRS}
    \subfloat[Cascade R-CNN Baseline w/ SCAResNet]{\includegraphics[scale=0.06]{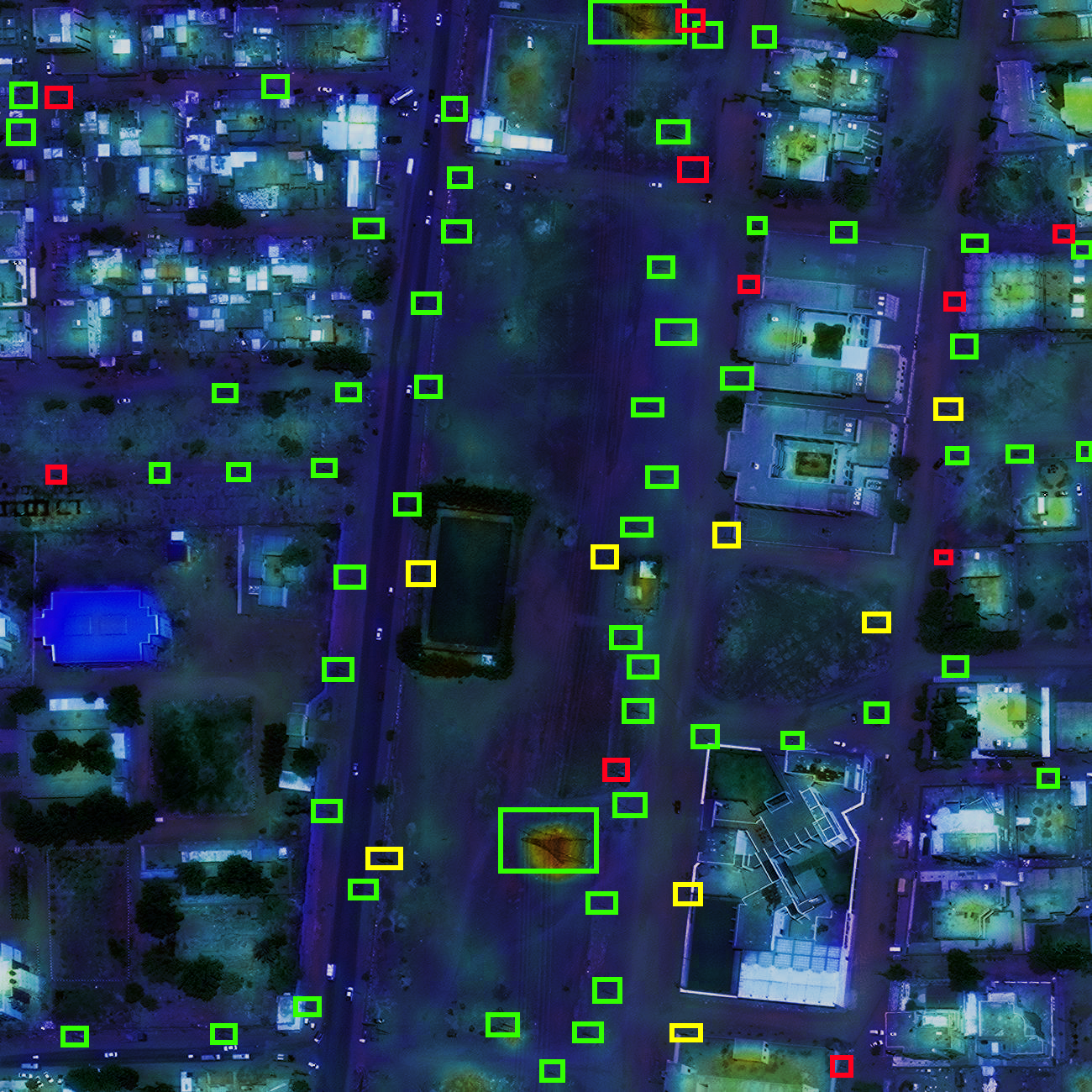}}%
    \label{CS}
    \caption{Comparison between RFLA-based as the baseline and SCAResNet-based baseline. The green box represents a correct detection, the red box represents a missed detection, and the yellow box represents a false detection.}
	\vspace{-2ex}
    \label{bassca}
\end{figure}

To better visualize the improvement in feature extraction of SCAResNet on transmission and distribution towers, we employed RFLA-based detectors as the baselines and baselines with SCAResNet to detect remote sensing images primarily containing distribution towers and transmission towers separately. As observed in Fig. \ref{bassca}, it is evident that the SCAResNet-enhanced RFLA-based detectors extract more abundant features and exhibit better detection performance for transmission and distribution towers.

\section {Conclusion}
We propose a backbone network, SCAResNet, designed explicitly for tiny object detection. We eliminate the conventional resizing operation during data preprocessing, as losing valuable information from the outset is detrimental for tiny objects like distribution towers. Our designed Positional-Encoding Multi-head CCA module enables learning more contextual features from images without losing information. The subsequent SPPRCSP module unifies feature maps from different sizes and scales into a consistent size and scale, enabling propagation without compromising accuracy while reducing parameters. SCAResNet achieves impressive detection results on ETDII Dataset.

\bibliographystyle{ieeetr}
\bibliography{ref}

\end{document}